# Facial Feature Tracking under Varying Facial Expressions and Face Poses based on Restricted Boltzmann Machines

Yue Wu     Zuoguan Wang     Qiang Ji

ECSE Department, Rensselaer Polytechnic Institute

{wuy9,wangz6,jiq}@rpi.edu

## Abstract

*Facial feature tracking is an active area in computer vision due to its relevance to many applications. It is a nontrivial task, since faces may have varying facial expressions, poses or occlusions. In this paper, we address this problem by proposing a face shape prior model that is constructed based on the Restricted Boltzmann Machines (RBM) and their variants. Specifically, we first construct a model based on Deep Belief Networks to capture the face shape variations due to varying facial expressions for near-frontal view. To handle pose variations, the frontal face shape prior model is incorporated into a 3-way RBM model that could capture the relationship between frontal face shapes and non-frontal face shapes. Finally, we introduce methods to systematically combine the face shape prior models with image measurements of facial feature points. Experiments on benchmark databases show that with the proposed method, facial feature points can be tracked robustly and accurately even if faces have significant facial expressions and poses.*

## 1. Introduction

Due to its relevance to many applications like human head pose estimation and facial expression recognition, facial feature tracking is an active area in computer vision. However, tracking facial feature points is challenging, since the face is non-rigid, and it can change its appearance and shape in unexpected ways. When faces have varying facial expressions and face poses, it is difficult to construct a prior model that could capture the large range of shape variations for facial feature tracking.

Restricted Boltzmann Machines (RBM) and their variants, such as Deep Belief Networks (DBNs) and 3-way RBM have proven to be versatile tools that could effectively solve some challenging computer vision tasks [5] [13] [17].

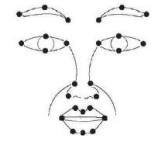

(a) 26 facial feature points that we track

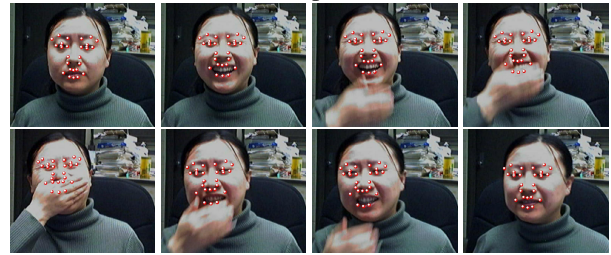

(b) one example sequence

Figure 1. Facial feature point tracking under expression variation and occlusion.

In recent years, these models have been used explicitly to handle the shape variations [17][5]. The nonlinearity embedded in RBM and its variants makes them more effective and efficient to represent the nonrigid deformations of objects compared to the linear methods. Their large number of hidden nodes and deep architectures also can impose sufficient constraints as well as enough degrees of freedoms into the representations of the target objects.

In this paper, we present a work that can effectively track facial feature points using face shape prior models that are constructed based on RBM. The facial feature tracker can track 26 facial feature points (Fig. 1 (a)) even if faces have different facial expressions, varying poses, or occlusion (Fig. 1 (b)). Unlike the previous works that track facial feature points independently or build a shape model to capture the variations of face shape or appearance regardless of the facial expressions and face poses, the proposed model could capture the distinctions as well as the variations of face shapes due to facial expression and pose change in a unified framework. Specifically, we first construct a model to capture the face shapes with different facial expressions for near frontal face based on Deep Belief Networks. Furthermore, to handle pose variations, we propose a model to capture the transition between the frontal face and non-frontal face with 3-way Restricted Boltzmann Machines. Finally, effective ways are proposed to combine the prior model with the image measurements of the facial feature points.

The paper is organized as follows: In section 2, we review the related research works. In section 3, we describe the face shape prior model that deals with expression variations based on DBNs (**FrontalRBM**). In section 4, we describe the face shape prior model that deals with both expression variations and pose variations based on 3-way RBM (**PoseRBM**). In section 5, we discuss how to combine the face shape prior model and measurements to generate final facial feature tracking results. In section 6, we show the experimental results. We conclude this paper in section 7.

## 2. Related work

### 2.1. Facial feature localization

Generally, facial feature localization algorithms in the literature can be classified into two categories: methods without face shape prior model and methods with face shape prior model. Methods without shape prior model track each facial feature point independently and ignore the prior knowledge about the face. As a result, they usually are susceptible to facial expression change, face poses change, occlusion etc. On the other hand, methods with shape prior model capture the dependence between facial feature points by explicitly modeling the general properties as well as the variations of facial shape or appearance. During testing, the shape can only deform within a certain range constrained by the model. Due to the limited space, we focus on the related methods with face shape prior model.

Typical methods with shape constraints are the Active Shape Model (ASM) [2] and Active Appearance Model (AAM) [12]. They build linear generative models to capture either shape variations or both shape and appearance variations of faces based on the Principle Component Analysis. More recently, in [20], facial feature points are detected independently based on the response of the support vector regressor. Then, the detection results are further constrained by a shape distribution embedded in a pre-trained Markov Random Field. Both [7] and [4] emphasize the facial components in their models. In [7], the shape variation of each facial component is modeled with a single Gaussian. In addition, the nonlinear relationship among facial components is represented as a pre-trained Gaussian Process Latent Variable model. In [4], Ding and Martinez propose a method to detect the facial feature points for each facial component with the use of subclass division and context information around features.

In real world situations, faces usually vary in facial expressions and poses. These natural movements make facial feature tracking even more difficult. To solve this problem, Tian and Cohn [9] propose a multi-state facial component model, where the state is selected by tracking a few control points. As a result, the accuracy of their method critically relies on how accurately and reliably the control points are tracked. Tong et al. [19] propose a model to capture the different states of facial components like mouth open and mouth closed. In addition, they project the frontal face to face with poses to handle the varying poses problem. However, during tracking, they need to dynamically and explicitly estimate the state of local components and switch between different models. In [3], Dantone and Van Gool take into account the pose variations and build sets of conditional regression forests on different poses. In [1], instead of using the parametric model, Belhumeur present methods to represent the face shape variations with non-parametric training data.

### 2.2. Restricted Boltzmann Machines based shape prior model

Recent works have shown the effectiveness of Restricted Boltzmann Machines and their variants in terms of representing objects' shapes. Due to the nonlinear nature embedded in these models, they are more suitable for capturing the variations of objects' shape, compared with the linear models like AAM [12] and ASM [2]. Recent research also shows that RBM and its variants can generalize well for unseen testing data. In [5], Eslami et al. propose a strong model of object shape based on Boltzmann Machines. Specifically, they build a Deep Belief Networks(DBNs)-like model but with only locally shared weights in the first hidden layer to represent the shape of horse and motorbikes. The sampling results from the model look realistic and have a good generalization. In [11], Luo et al. train DBNs for detecting the facial components like the mouth, eyes, nose and eyebrows. In their model, the input are the image patches; they pre-train DBNs using layer-wise RBM and then use the labels to fine-tune the model for classification with logistic regression. RBM also has been applied to model finger motion patterns and human body pose, and used as the prior to learn features [23] and decoders [22]. In [17], Taylor et al. propose to use a new prior model called implicit mixture of Conditional Restricted Boltzmann Machines to capture the human poses and motions (imRBM). The mixture nature of imRBM makes it possible to learn a single model to represent the human poses and motions under different activities such as walking, running, etc. In [8], Susskind et al. build a generative model to capture the joint probability of the facial action units, identities, and binary image patch data using the DBNs. Then, based on this model, they can generate realistic binary face patches displaying specific identities and facial actions.

## 3. Prior model for face shapes with different facial expressions (FrontalRBM)

Although the appearance of the human face varies from individual to individual, the spatial relationship among facial feature points is similar for a given facial expression. As can be seen from Fig. 2, there exist patterns for human face shapes, but these patterns depend on the facial expressions. To capture these patterns, we propose a face shape prior model based on Deep Belief Networks which we call FornltalRBM in this paper. It's structure is shown as part I in Fig. 3.

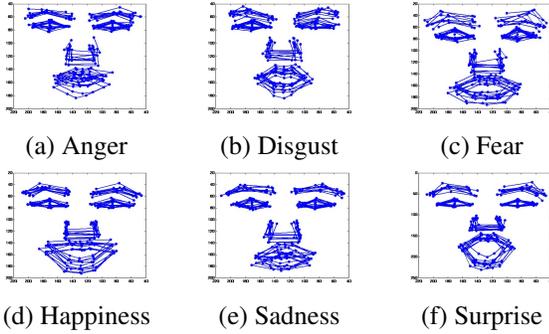

(a) Anger  (b) Disgust  (c) Fear

(d) Happiness  (e) Sadness  (f) Surprise

Figure 2. Facial feature locations of 10 subjects with different facial expressions

Deep belief networks (DBNs) are well known as an approach to automatically extract effective multi-level features from data. Moreover, because of its ability to model the distribution of observations, recently DBNs are exploited to be used as the shape prior models [5] [17].

The building block of DBNs is Restricted Boltzmann Machines (RBM), which represent the probabilistic density function of the observed variable $\mathbf{x}$ as:

$$p_\eta(\mathbf{x}) = \frac{1}{Z} \sum_{\mathbf{h}} e^{-E_\eta(\mathbf{x},\mathbf{h})}, \quad (1)$$

where $Z$ is the normalizing constant, and $\mathbf{h} \in \{0,1\}^{\mathcal{H}}$ are binary hidden variables. The pdf is defined in terms of the joint energy function over $\mathbf{x}$ and $\mathbf{h}$, as:

$$-E_\eta(\mathbf{x},\mathbf{h}) = \sum_i \mathbf{b}_i \mathbf{x}_i + \sum_{i,j} \mathbf{w}_{ij} \mathbf{x}_i \mathbf{h}_j + \sum_j \mathbf{c}_j \mathbf{h}_j, \quad (2)$$

where $\mathbf{w}_{ij}$ is the interaction strength between the hidden node $\mathbf{h}_j$ and the visible node $\mathbf{x}_i$. $\mathbf{b}$ and $\mathbf{c}$ are the biases for the visible layer and hidden layer. The parameters in this model, $(\mathbf{w}, \mathbf{c}, \mathbf{b})$, are collectively represented as $\eta$.

In the application of facial feature points tracking, our goal is to use two-layer DBNs to explicitly capture the face shape patterns under different facial expressions. In this case, the observation nodes are the coordinates of facial feature locations, normalized according to the locations of

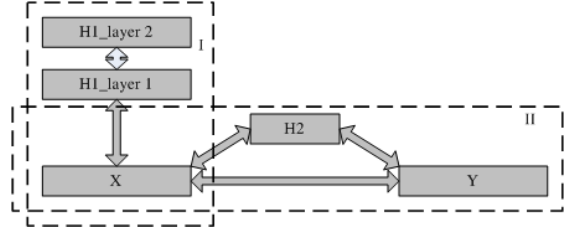

Part I: Frontal face shape prior with different expressions

Part II: Transfer to different poses

Figure 3. Face shape prior model based on the combination of the DBNs and 3-way RBM

eyes, denoted as $x = [p_{1,x}, p_{1,y}, p_{2,x}, p_{2,y}...p_{26,x}, p_{26,y}]^t$. Since the observations are continuous, we use the *Gaussian-Bernoulli Restricted Boltzmann Machine* (GB-RBM) [14] in the bottom layer. Similar to standard RBM, the energy function of GB-RBM is defined as:

$$E_\eta(\mathbf{x},\mathbf{h}) = \sum_i \frac{(\mathbf{x}_i - \mathbf{b}_i)^2}{2} - \sum_{i,j} \mathbf{w}_{ij} \mathbf{x}_i \mathbf{h}_j - \sum_j \mathbf{c}_j \mathbf{h}_j. \quad (3)$$

Direct maximum likelihood parameter estimation of RBM/GB-RBM is intractable for high dimensional models due to the normalizing factor $Z$, so we use Contrastive Divergence algorithm (CD). The multi-layer DBNs are trained in a layer-wise manner so that training is relatively efficient. Please refer to [14] and [6] for more detailed discussion about GB-RBM and CD algorithm.

## 4. Prior model for face shapes with varying facial expressions and poses (PoseRBM)

It is nature to think of extending the DBNs described in the last section to capture shape variations with both facial expression and pose changes. However, this raises a problem when the number of possible poses increases, and we cannot expect the DBNs to learn all possible face shape patterns with all poses well unless a large number of hidden nodes are used, which would significantly increase the number of required training data. To alleviate this problem, we propose a novel face shape prior model as shown in Fig. 3, where $x$ represents the locations of facial feature points for frontal face when subjects show different facial expressions, and $y$ represents the corresponding locations of facial feature points for non-frontal face under the same facial expression. $H_1$ and $H_2$ are two sets of hidden nodes. In Fig. 3, the face shape prior model is factorized into two parts. In part I, the two layer DBNs model captures the shape patterns of the frontal face under varying facial expressions as discussed in the previous section. In part II, the 3-way RBM model captures the transition between the facial feature locations for frontal face and corresponding non-frontal face. The two parts of the model in Fig. 3 can be trained separately.

Part II is realized by the 3-way RBM, which have been shown to capture the corresponding relationship between two images [13]. To reduce model parameters, we use factorized 3-way RBM, whose energy function can be written as:

$$-E(\mathbf{x}, \mathbf{h}, \mathbf{y}) = \sum_f (\sum_i \mathbf{x}_i \mathbf{w}_{if}^x)(\sum_j \mathbf{y}_j \mathbf{w}_{jf}^y)(\sum_k \mathbf{h}_k \mathbf{w}_{kf}^h) \\ - \sum_i \frac{1}{2}(\mathbf{x}_i - \mathbf{w}_i^x)^2 - \sum_j \frac{1}{2}(\mathbf{x}_j - \mathbf{w}_j^y)^2 + \sum_k \mathbf{w}_k^h \mathbf{h}_k, \quad (4)$$

where $\mathbf{h}$ is corresponding to $H_2$ in Fig. 3. The parameters $\mathbf{w}_{if}^x$, $\mathbf{w}_{jf}^y$ and $\mathbf{w}_{kf}^h$ describe the interaction among variables, and $\mathbf{w}_i^x$, $\mathbf{w}_j^y$ and $\mathbf{w}_k^h$ are the bias for $\mathbf{x}$, $\mathbf{y}$ and $\mathbf{h}$. In the previous work [13], the factorized 3-way RBM is learned by maximizing the conditional likelihood $p(y|x)$. In this work, to facilitate the combination of measurements and prior model, the parameters are optimized by maximizing the joint likelihood $p(x, y)$. The derivative [13] of log-likelihood $L(x, y|\mathbf{w})$ can be written as

$$\frac{\partial L}{\partial \mathbf{w}} = \langle \frac{\partial E}{\partial \mathbf{w}} \rangle_{p(\mathbf{x},\mathbf{h},\mathbf{y}|\mathbf{w})} - \langle \frac{\partial E}{\partial \mathbf{w}} \rangle_{p(\mathbf{h}|\mathbf{x},\mathbf{y},\mathbf{w})}, \quad (5)$$

where $\langle . \rangle_p$ represents expectation over $p$. It is difficult to directly calculate equation 5. We approximate it with contrastive divergency [6], which can be realized with Gibbs sampling. Since given any two variables in $\mathbf{x}$, $\mathbf{h}$ and $\mathbf{y}$, the elements in the remaining variable are independent, we can readily implement the Gibbs sampling. Given $\mathbf{x}$ and $\mathbf{y}$, the hidden variable $\mathbf{h}$ can be sampled from:

$$p(\mathbf{h}_k = 1|\mathbf{x}, \mathbf{y}) = \sigma(\sum_f \mathbf{w}_{kf}^h (\sum_i \mathbf{x}_i \mathbf{w}_{if}^x)(\sum_j \mathbf{y}_j \mathbf{w}_{jf}^y) + \mathbf{w}_k^h), \quad (6)$$

where $\sigma$ represents the sigmoid function $\sigma(x) = 1/(1 + e^{-x})$. The element of $\mathbf{x}$ or $\mathbf{y}$ given the remaining variables follows a Gaussian distribution with unit variance and mean

$$\mu_{\mathbf{x}_i} = \sum_f \mathbf{w}_{if}^x (\sum_k \mathbf{h}_k \mathbf{w}_{kf}^h)(\sum_j \mathbf{y}_j \mathbf{w}_{jf}^y) + \mathbf{w}_i^x, \quad (7)$$

$$\mu_{\mathbf{y}_j} = \sum_f \mathbf{w}_{jf}^y (\sum_i \mathbf{x}_i \mathbf{w}_{if}^x)(\sum_k \mathbf{h}_k \mathbf{w}_{kf}^h) + \mathbf{w}_j^y. \quad (8)$$

With the derivative of $\frac{\partial L}{\partial \mathbf{w}}$, the parameters $\mathbf{w}$ are optimized with stochastic gradient.

## 5. Facial feature tracking based on face shape prior model

Facial feature tracking accuracy and robustness can be improved by incorporating the face shape prior model. Assume the true facial feature locations we want to infer are denoted as $X^*$ and their measurements are represented as $X_m$, facial feature tracking can be regarded as an optimization problem:

$$X^* = \arg\max_X P(X|X_m) \\ = \arg\max_X P(X_m|X)P(X) \quad (9)$$

Normally, $P(X_m|X)$ is modeled by the multivariate Gaussian distribution:

$$P(X_m|X) = \frac{1}{(2\pi)^{\frac{k}{2}}|\Sigma_l|^{-\frac{1}{2}}} e^{-\frac{1}{2}(X-X_m)^t \Sigma_l^{-1}(X-X_m)}, \quad (10)$$

where $\Sigma_l$ is the covariance matrix that can be estimated from the training data.

It is difficult to analytically formulate the prior probability $P(X)$ from the learned models described in previous two sections, we hence propose to estimate the prior probability numerically via sampling. Since facial feature points vector is of high dimensions (52 in this case), to globally produce enough samples to cover such a large parameter space is computationally infeasible. We propose to perform local sampling instead. During tracking, we only need to estimate the local prior probability around $X_m$.

If the face shape only varies due to facial expression change, we use FrontalRBM discussed in section 3 as shape prior. In this case, with the measurement as the initialization of nodes $x$, MCMC sampling method can generate samples that fit the model. If the face shape varies due to both facial expression and pose change, we use PoseRBM described in section 4 as prior. In this case, we first use MCMC sampling to sample a few times through part II of the model to generate $x$ with the measurement as the initialization of $y$ in Fig. 3. Second, we sample $x$ a few times through part I of the model. With the sampling results of $x$ from part I and the hidden nodes $H_2$ that generated in the first step, we can sample $y$ that fits the PoseRBM prior model.

To combine the sampling results with the measurement, we propose two approaches shown in the following.

**Gaussian assumption:** With the strong assumption that the local prior probability is multivariate Gaussian distribution, we could estimate the prior probability by calculating the mean vector $\mu_p$ and covariance matrix $\Sigma_p$ from the samples. Since the likelihood probability is also multivariate Gaussian (Eq. 10), $X^*$ can be calculated analytically:

$$X^* = (\Sigma_l^{-1} + \Sigma_p^{-1})^{-1}(\Sigma_p^{-1}\mu_p + \Sigma_l^{-1}X_m) \quad (11)$$

**Kernel Density Function:** Given the samples, a typical non-parametric way to estimate the probability is using Kernel Density Estimation method. When using multivariate Gaussian as the kernel function, the prior probability becomes:

$$p(X) = \sum_d \frac{1}{(2\pi)^{\frac{k}{2}}|\Sigma_k|^{-\frac{1}{2}}} e^{-\frac{1}{2}(X-X_d)^t \Sigma_k^{-1}(X-X_d)}, \quad (12)$$

where $X_d$ indicates one sample with $d$ as the index, and parameter $\Sigma_k$ is estimated from the samples. Based on Eq. 9,10,12, we can use typical optimization method to estimate the optimal value $X^*$, given $X_m$, $X_d$, $\Sigma_k$ and $\Sigma_l$.

## 6. Experimental results

### 6.1. Facial feature tracking under different facial expressions

In this section, we test the FrontalRBM model described in section 3 that could track facial feature points for near-frontal face with different facial expressions. We show the experiments using synthetic data, sequences from the extended Cohn-Kanade database (CK+) [10], the MMI facial expression database [15], the American Sign Language (ASL) database [24] and the ISL Facial Expression database [18].

**Experiments on synthetic data:** Fig. 4 (a) and (c) are faces with outlier (left eyebrow tip) and corrupted points on the left half face. Fig. 4 (b) and (d) are the results after correction. FrontalRBM shows strong power as a face shape prior model.

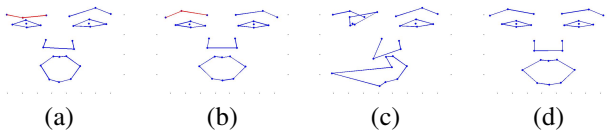

(a)　　　　(b)　　　　(c)　　　　(d)

Figure 4. Performance of FrontalRBM based on synthetic data. (a) face with outlier (left eyebrow tip); (b) Correction of (a); (c) face with corrupted points on left half face; (d) correction of (c).

**Experiments on CK+ database:** The CK+ database contains facial behavior videos of 123 subjects showing 7 basic facial expressions including anger, disgust, fear, happiness, sadness, surprise, and contempt. Since the number of sequences for contempt is much smaller than the other facial expressions, we exclude the sequences with contempt facial expression. In total, there are 282 sequences. We manually label the first, middle (onset), and last (apex) frames for each sequence as ground truth and use the tracking results generated using AAM [12] which are provided by the database as measurements. The FrontalRBM model is trained by the manually labeled data. It is important to notice that data with neutral facial expression is included as the first frame of each sequences.

To evaluate the performance of the tracking algorithms, we use the distance error metric for point $i$ at frame $j$ as follows:

$$Error_{i,j} = \frac{\|P_{i,j} - \hat{P}_{i,j}\|_2}{D_I(j)}, \quad (13)$$

where $D_I(j)$ is the interocular distance measured at frame $j$, $\hat{P}_{i,j}$ is the tracked point $i$ at frame $j$, and $P_{i,j}$ is the manually labeled ground truth.

The experimental results on CK+ database are illustrated in Table 1, Fig. 5, and Fig 6. From Table 1, we could see that by using the proposed model and leave-one-subject-out cross validation, the overall tracking errors decrease by 16.88% and 12.24% when using gaussian assumption (proposed method 1) and kernel density estimation (proposed method 2) to combine the model and the measurement as described in section 5. Fig. 5 compares the tracking error for different facial expressions. We can see that the proposed method can decrease the tracking error for all the facial expressions and the performances are similar. Fig. 6 shows the tracking results for one sequence. In the original sequence (Fig. 6 (a)), the subject shows relatively neutral facial expression at the beginning and then happy facial expression after the forth frame. Compared with the baseline method, the proposed method could decrease the tracking error for each frame.

|  | Eyebrow | Eye | Nose | Mouth | Overall |
|---|---|---|---|---|---|
| baseline method [12] | 8.7367 | 4.5367 | 7.7240 | 3.9704 | 5.8220 |
| proposed method 1 | 7.4111 | 3.6214 | 5.2871 | 3.9046 | 4.8394 |
| Proposed method 2 | 7.4472 | 3.8019 | 5.5654 | 4.4350 | 5.1092 |
| Overall improvement | Proposed method 1: 16.88%. Proposed method 2: 12.24% | | | | |

Table 1. Experimental results on CK+ database. Baseline method: AAM [12]. Proposed method 1: FrontalRBM, combined under Gaussian assumption. Proposed method 2: FrontalRBM, combined using KDE.

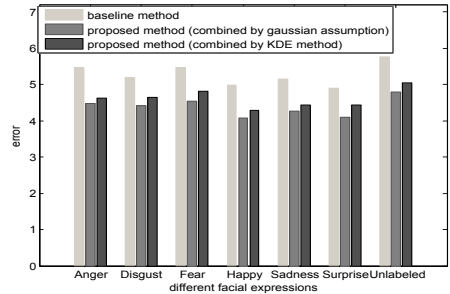

Figure 5. Tracking errors for different facial expressions on CK+ database. Baseline method: AAM [12]. Proposed method: FrontalRBM.

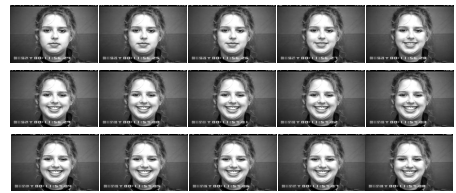

(a) Sequence

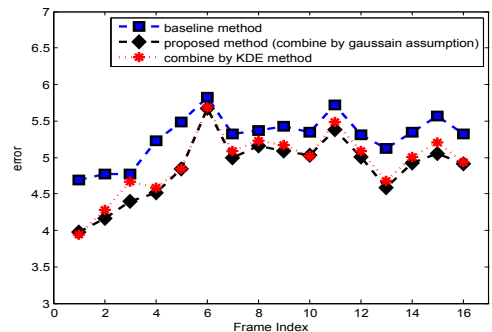

(b) Facial feature tracking results.

Figure 6. Tracking error on one sample sequence from CK+ database. Baseline method: AAM [12]. Proposed method: FrontalRBM.

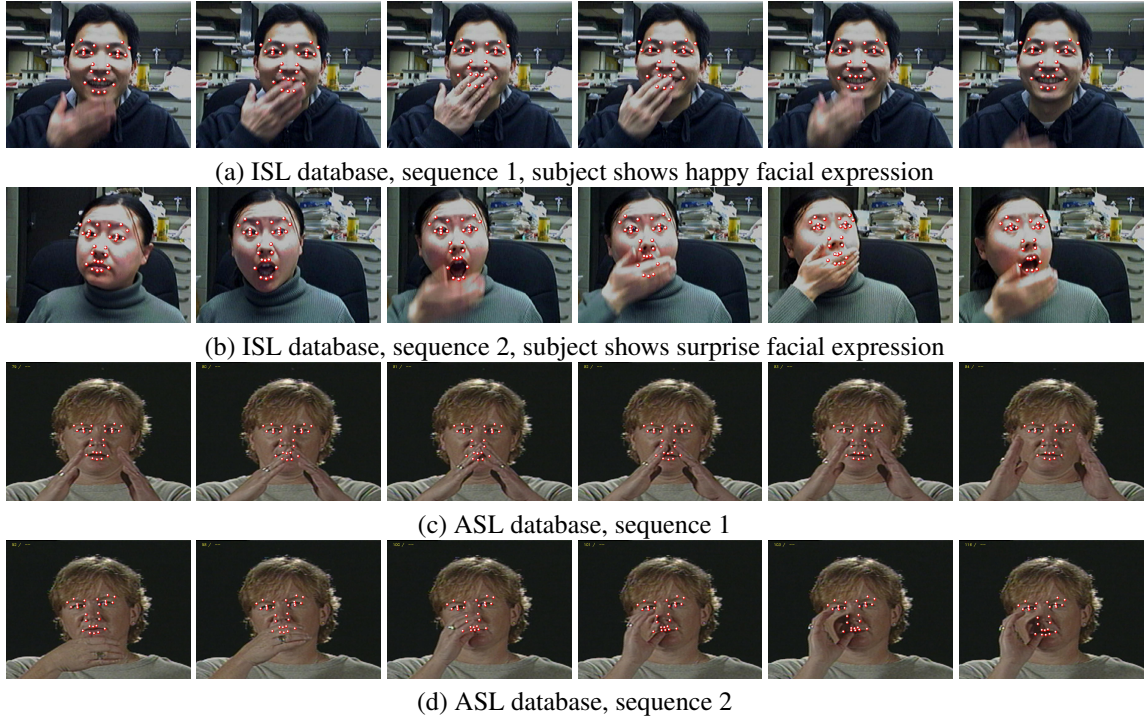

Figure 7. Facial feature tracking results on ISL occlusion database and ASL database using proposed method (FrontalRBM).

(a) ISL database, sequence 1, subject shows happy facial expression

(b) ISL database, sequence 2, subject shows surprise facial expression

(c) ASL database, sequence 1

(d) ASL database, sequence 2

**Experiments on MMI database:** In our experiments on MMI database [15], there are 196 sequences of 27 subjects with 6 basic facial expressions. Some subjects may wear glasses. We generate the tracking results using the switch model [19] as the measurements. We manually label the onset and apex frames for each sequence with facial expressions and the first frame as neutral expression. With the leave-one-subject-out strategy, the results are shown in Table 2. By incorporating the frontalRBM as face shape prior model, the overall errors decrease by 16.50% and 15.13% when using Gassuain assumption and KDE to combine the measurement and the model. Our result is comparable to the state of art research [21], which reports an average detection error of 5.3 on 400 images selected from not only the MMI database but also the FERET database [16].

|  | Eyebrow | Eye | Nose | Mouth | Overall |
|---|---|---|---|---|---|
| baseline method [19] | 8.3677 | 4.3418 | 7.0194 | 7.3861 | 6.6195 |
| proposed method 1 | 6.6433 | 3.3132 | 5.2956 | 7.0210 | 5.5275 |
| Proposed method 2 | 6.7289 | 3.4146 | 5.5356 | 7.0288 | 5.6178 |
| Overall improvement | Proposed method 1: 16.50%. Proposed method 2: 15.13% | | | | |

Table 2. Experimental results on MMI database. Baseline method: Switch Model [19]. Proposed method 1: FrontalRBM, combined under Gaussian assumption. Proposed method 2: FrontalRBM, combined based on KDE.

**Dealing with occlusion:** In real world situations, the facial features may be occluded by other objects. To test the proposed method under occlusion, we perform tracking on the ISL occlusion facial expression database [18] and some sequence with occlusion from American Sign Language (ASL) database [24]. Specifically, the ISL occlusion database consists of 10 sequences of 2 subjects showing happy and surprised facial expressions with near-frontal pose. The face shape prior model is trained using the manually labeled data on CK+ database. To perform tracking using the proposed method, we use the simple Kalman Filter, local image intensity patch as feature and manually label the facial features on the first frame. For each frame, the tracked points are constrained by the FrontalRBM and use Gaussian assumption as the combination strategy. The experimental results are shown in Fig. 1 (b) and Fig. 7. The proposed tracker can correctly track the facial features under occlusion.

### 6.2. Facial feature tracking under different facial expressions and varying face poses

In this section, we report the test of the model proposed in section 4 (**PoseRBM**) that could track facial features when faces have simultaneous expression and pose changes.

**Experiments using synthetic data:** Fig. 8 shows the experimental results of PoseRBM based on synthetic data. Face shapes here are all with 22.5 degrees to the left. Similar to the performance shown in Fig. 4, PoseRBM as a face shape prior model can correct the outliers and even the corrupted points on half of the face.

**Experiments on ISL multi-view facial expression database [18]:** Overall, there are 40 sequences of 8 subjects showing happy and surprised facial expressions under varying face poses. Some face poses are very significant. To train the model illustrated in Fig. 3, we can train part I and part II separately. Part I is trained using the manually labeled data on CK+ database as described in the previous subsection. In order to train part II, we need the corresponding facial feature points for frontal face and non-frontal face with the same expression. As far as our knowledge, there is no such database with large number of real corresponding images available to the public, so we project the frontal facial feature points in CK+ database to specific poses based on general z coordinates of the feature points. Here, to train the PoseRBM model, we use the frontal face and generated non-frontal face with 22.5 degrees to the left and right. Although it is trained with data from three poses, PoseRBM can adapt to deal with the poses under moderate angles(less than 50 degrees) where full view of all facial features is available. Such model flexibility comes from three aspects. First, RBM can generalize across prior information included in the training data, which is demonstrated by the fact that RBM can generate samples closed to the training data, but never appeared [5]. Second, PoseRBM factorizes the modeling into two steps, i.e., modeling expressions for frontal face and rotation to different poses. This automatically adapts the expressions in frontal face to other poses, removing the necessary of modeling expressions under each pose. Third, the in-plane rotation is excluded by normalizing the face shape according to the locations of eyes.

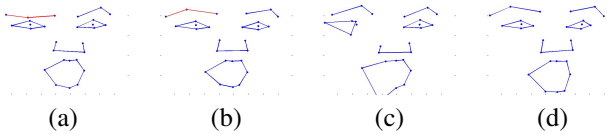

(a)         (b)         (c)         (d)

Figure 8. Performance of PoseRBM based on synthetic data. (a) face with outlier (left eyebrow tip); (b) Correction of (a); (c) face with corrupted points on left half face; (d) correction of (c).

Tracking is performed by incorporating the PoseRBM model, simple Kalman Filter, local image patch as features, and manually labeled facial features on the first frame. We use the switch model [19] as the baseline method. In order to generate quantitative results, we manually label every 5 frames and some frames with significant appearance change for every sequence.

The experimental results are shown in Table 3, Fig. 9 and Fig. 10. Table 3 shows the overall tracking error. It can be seen that the proposed method decreases the tracking error for both happy and surprised facial expressions. Fig. 9 compares the error distributions of facial feature tracking for happy and surprise expressions, respectively. It can be seen that higher percentages of frames have lower tracking errors when using the proposed method than the baseline method. Fig. 10 shows some tracking results when subjects turn right and left with different facial expressions. The tracking results are accurate if the pose angle is relatively moderate. The tracking error may increase if the subject shows simultaneous extreme face poses and facial expressions(last frame in 10(e)(f)).

|                 | Happy  | Surprise |
|-----------------|--------|----------|
| Baseline method | 8.6370 | 8.6396   |
| Proposed method | 6.3539 | 7.2009   |

Table 3. Experimental results on ISL multi-view facial expression database. Baseline method: switch model [19]. Proposed method: PoseRBM.

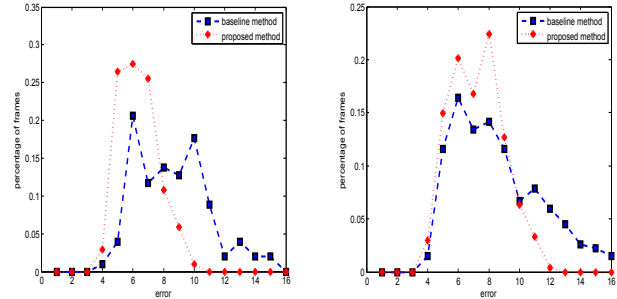

(a) Happy facial expressions    (b) Surprised facial expressions

Figure 9. Error distributions of facial feature tracking. Baseline method: switch model [19]. Proposed method: PoseRBM.

## 7. Conclusions and future work

In this paper, we introduce methods to construct a face shape prior model based on Restricted Boltzmann Machines and their variants to improve the accuracy and robustness of facial feature tracking under simultaneous pose and expression variations. Specifically, we first introduce a face shape prior model to capture the face shape patterns under varying facial expressions for near-frontal face based on Deep Belief Networks. We then extend the frontal face prior model by a 3-way RBM to capture face shape patterns under simultaneous expression and pose variation. Finally, we introduce methods to systematically combine the face prior models with image measurements of facial feature points to perform facial feature point tracking. Experimental comparisons with state of the art methods on benchmark data sets show the improved performance of the proposed methods even when faces have varying facial expressions, poses, and occlusion. In the future, we will perform further validation of our methods and capture the dynamic relationship between frames.

**Acknowledgements:** This work is supported in part by a grant from US Army Research office (W911NF-12-C-0017).

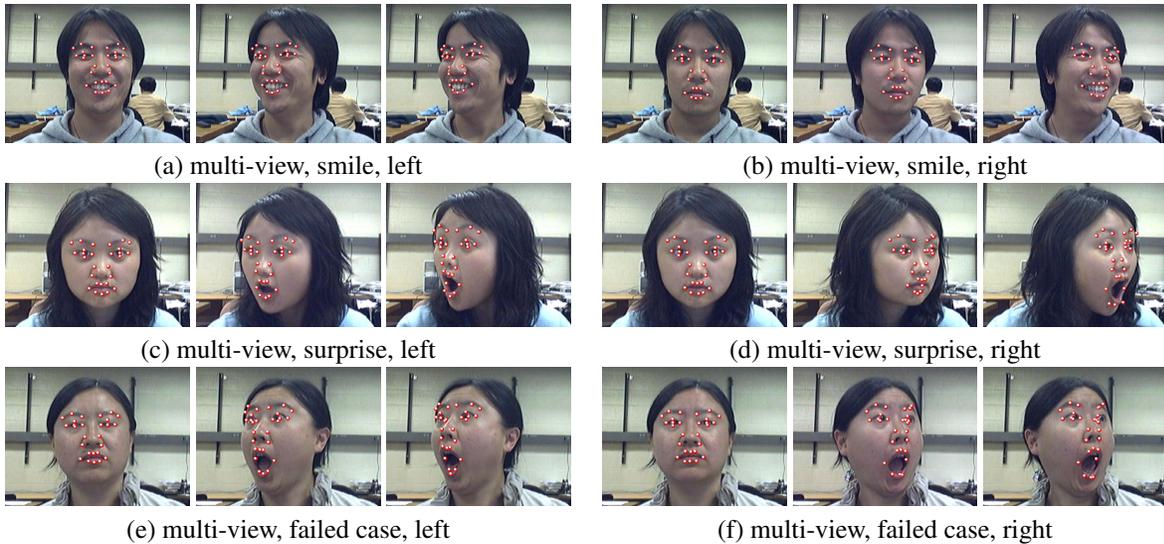

Figure 10. Facial feature tracking results on ISL multi-view facial expression database using the proposed method (PoseRBM).

(a) multi-view, smile, left
(b) multi-view, smile, right
(c) multi-view, surprise, left
(d) multi-view, surprise, right
(e) multi-view, failed case, left
(f) multi-view, failed case, right